\documentclass{article} 
\usepackage{iclr2027_conference,times}

\usepackage{amsmath,amsfonts,bm}

\def\eqref#1{equation~\ref{#1}}

\def\1{\bm{1}}

\DeclareMathAlphabet{\mathsfit}{\encodingdefault}{\sfdefault}{m}{sl}
\SetMathAlphabet{\mathsfit}{bold}{\encodingdefault}{\sfdefault}{bx}{n}

\usepackage{hyperref}
\usepackage{url}
\usepackage{booktabs}       
\usepackage{amsfonts}       
\usepackage{nicefrac}       
\usepackage{microtype}      
\usepackage{xcolor}         
\usepackage{graphicx}
\usepackage{subcaption}
\usepackage[ruled,vlined,linesnumbered]{algorithm2e}
\usepackage{multirow}
\usepackage{wrapfig}
\usepackage{amsmath}
\usepackage{amsthm}

\newcommand{\std}[1]{{\scriptsize(#1)}}

\title{Query-based Submodular Data Selection for Efficient Instruction Tuning}

\author{Zhenghao Zhao\textsuperscript{1},
    Yasutomo Kawanishi\textsuperscript{2},
    Huihuo Zheng\textsuperscript{3},
    Michael E. Papka\textsuperscript{1},
    Yan Yan\textsuperscript{1} \\
    \textsuperscript{1}University of Illinois Chicago \quad
    \textsuperscript{2}Ritsumeikan University \quad
    \textsuperscript{3}Argonne National Laboratory
}

\iclrfinalcopy 
\begin{document}

\maketitle
\lhead{Preprint.}

\begin{abstract}
As large language models are fine-tuned for increasingly specialized tasks,
their performance depends more on which training examples are selected than on
how many.
Given a large heterogeneous training pool and a small target set specifying the desired capability, targeted data selection aims to find a compact subset that maximizes downstream performance.
Existing methods score each candidate independently against the target and
select the top-ranked examples, yet this point-wise strategy ignores inter-example interactions: at tight budgets the selected subset is dominated by near-redundant points that waste capacity on overlapping information.
We propose \textbf{Q}uery-based \textbf{S}ubmodular functions with \textbf{G}radient-guided \textbf{S}ynthesis (QSGS), which formulates targeted selection as a submodular graph cut over a gradient similarity graph.
A modular relevance term pulls selected examples toward the target task, while a pairwise dispersion penalty suppresses intra-subset redundancy. 
The resulting objective balances target relevance and intra-subset redundancy using the same formulation across different selection budgets. Empirically, the benefit of diversity-aware selection becomes increasingly pronounced as the data budget decreases. 
To further close coverage gaps that no selection from existing data can bridge, we introduce a gradient-guided synthesis step that identifies under-covered target directions via residual subspace analysis and generates synthetic examples to fill them.
Experiments on MMLU, BBH, and TydiQA across Llama-2, Llama-3, and Mistral, show that QSGS at a 0.5\% budget matches or exceeds full-data performance on Mistral-7B, outperforms the strongest baseline across all benchmarks, and yields increasing gains as the budget shrinks.
\end{abstract}    
\section{Introduction}
\label{sec:intro}

As large language models~\citep{touvron2023llama,grattafiori2024llama,jiang2023mistral} are deployed in increasingly specialized roles, from multilingual question answering~\citep{tydiqa,lewis2020mlqa} to complex multi-step reasoning\citep{wei2022chain,wang2022self,yao2023tree}, their effectiveness hinges less on the scale of pretraining than on the quality and composition of the data used to fine-tune them~\citep{longpre2023flan,zhou2023lima}.
LIMA~\citep{zhou2023lima} demonstrates that a  few carefully chosen examples can rival models trained on orders of magnitude more data, underscoring a core insight: {which} examples a model trains on matters far more than {how many}.
This motivates a principled data selection problem: given a large, heterogeneous training pool and a small set of target demonstrations that specify a desired capability, how should we select a compact subset to maximize downstream performance?

A growing body of work has tackled data selection for instruction tuning from diverse angles. Quality-based methods use heuristic scores such as instruction-following difficulty~\citep{li2024quantity} or LLM-judged complexity and quality~\citep{chen2023alpagasus,liu2023makes,du2023mods} to filter or rank examples, while distributional approaches match the selected subset to a reference distribution~\citep{xie2023data}.
Data pruning techniques identify and remove uninformative samples via gradient-based importance estimates~\citep{tan2023data}, and coreset methods enforce diversity through clustering or submodular optimization~\citep{wei2015submodularity,zhang2025tagcos}.
Among these, LESS~\citep{xia2024less} achieves strong targeted performance by estimating per-example influence via gradient similarity and selecting the top-$K$ highest-scoring candidates.

\sloppy
However, all these methods score each candidate independently against the target set and select the top-ranked examples.
This point-wise strategy ignores how selected examples interact with one another: as the budget shrinks, top-ranked candidates increasingly share similar gradient directions, so the selected subset wastes its limited capacity on redundant signals instead of covering complementary aspects of the target task.
The tighter the budget, the larger the marginal cost of each redundant selection, yet the scoring criterion remains oblivious to intra-subset overlap.
This analysis motivates a selection objective that jointly accounts for target relevance and inter-example redundancy, with diversity pressure that adapts to the budget size.

\begin{wrapfigure}{r}{0.5\textwidth}
    \centering
    \vspace{-0.4cm}
    \includegraphics[trim=10 20 105 30,clip,width=1\linewidth]{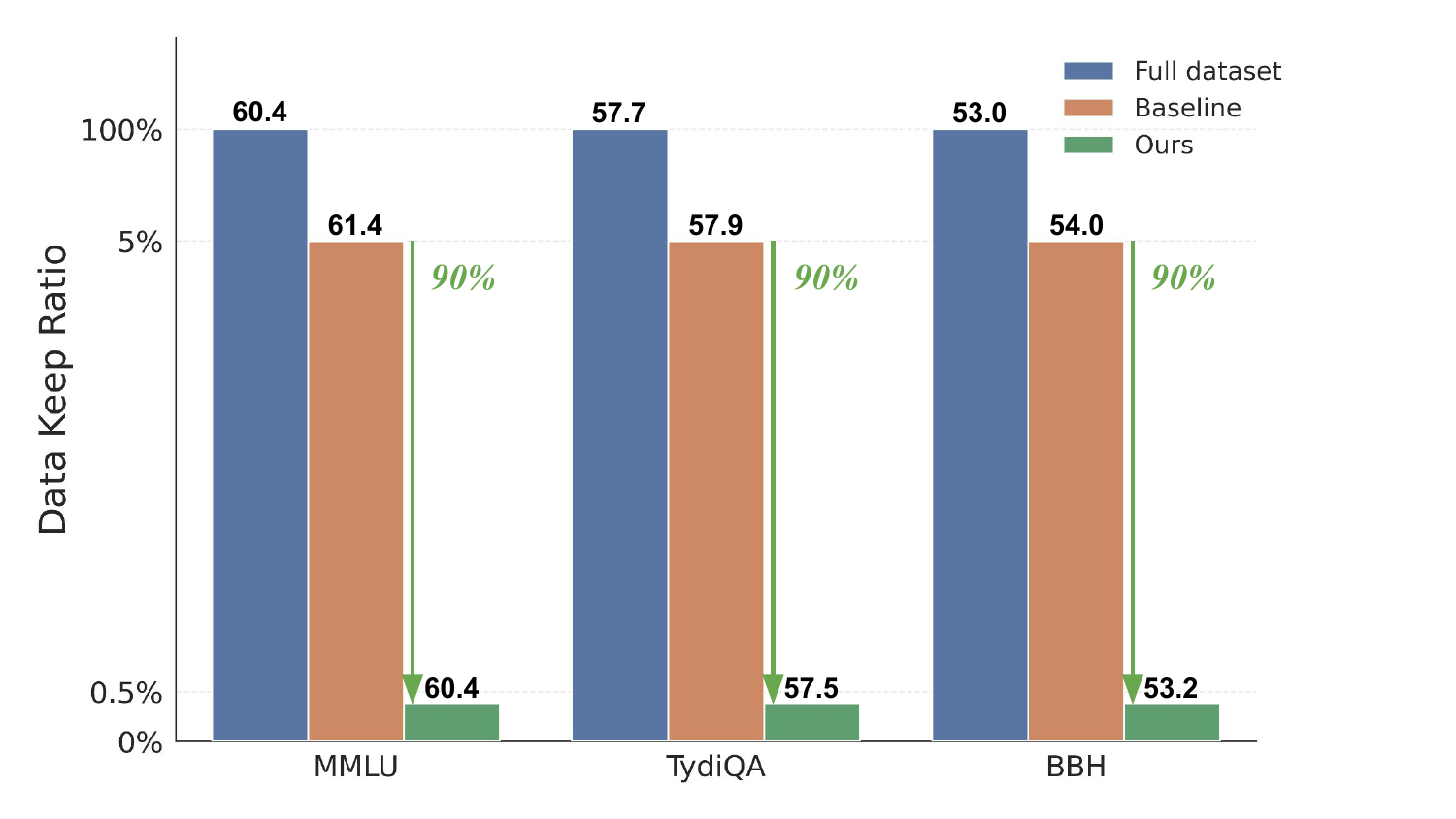}
    \caption{\textbf{Losses dataset compression with QSGS framework.} On all three benchmarks, our method achieves full set performance with using only 0.5\% of the data. The data keep ratio of ours is 90\% less than the SOTA method~\citep{xia2024less}.}
    \label{fig:teaser}
    \vspace{-10pt}
\end{wrapfigure}

\sloppy
We propose \textbf{Q}uery-based \textbf{S}ubmodular functions with \textbf{G}radient-guided \textbf{S}ynthesis (QSGS), a unified coreset construction framework that operates in three tightly coupled stages built on a shared gradient representation.
First, low-rank gradient features are computed from early-stage checkpoints of a warm-up model, yielding both per-example influence scores against the target set and compact gradient embeddings for pairwise similarity computation.
Second, a target-aware submodular graph cut selects examples by jointly maximizing relevance to the target task and minimizing intra-subset redundancy in gradient space.
This subset-level objective is useful under tight budgets, where each redundant selection consumes a larger fraction of the limited capacity and displaces a potentially complementary example.
Third, a gradient-guided synthesis step identifies target gradient directions that are uncovered by the selected subset and generates synthetic examples to fill the residual gaps.

We evaluate QSGS at an extreme 0.5\% data budget, which contains only 1{,}350 out of 270{,}000 examples across diverse benchmarks and model families. QSGS consistently outperforms LESS across the evaluated settings, notably matching full-data MMLU performance, as shown in Figure~\ref{fig:teaser}. Importantly, its advantage becomes more pronounced as the available data decreases, demonstrating strong robustness in highly data-scarce regimes.
Our ablation studies show that the graph-cut objective accounts for the majority of the performance improvement, with gradient-guided synthesis providing additional gains, while maintaining a selection overhead below 0.1\% of the total pipeline cost.

Our contributions are as follows:
\begin{itemize}
    \item We propose QSGS, a unified coreset selection framework for LLM instruction tuning that first computes gradient-based influence scores, then selects samples based on influence scores, and fills residual coverage gaps through gradient-guided data synthesis.
    \item We design a target-aware submodular graph cut objective that jointly maximizes candidate--target relevance and minimizes intra-subset redundancy in gradient space. We also introduce a gradient-guided synthesis method that identifies uncovered target gradient directions via residual subspace analysis and generates targeted synthetic examples to fill the gaps.
    \item Experiments across three benchmarks (MMLU, TydiQA, BBH) and three model families (Llama-2, Llama-3, Mistral) show that QSGS at 0.5\% budget matches or exceeds full-data performance on Mistral-7B, while the gap widens as the budget shrinks. 
\end{itemize}

\section{Related Work}
\label{sec:related}

\noindent\textbf{Instruction tuning and data curation.}
Instruction tuning~\citep{ouyang2022training} aligns pretrained language models with downstream tasks by fine-tuning on instruction--response pairs~\citep{longpre2023flan,wei2021finetuned,chung2024scaling}.
As training pools grow increasingly heterogeneous, the composition of the fine-tuning mixture has been shown to matter more than its sheer size~\citep{zhou2023lima,chen2023alpagasus}: LIMA~\citep{zhou2023lima} demonstrates that as few as 1{,}000 carefully curated examples can match the performance of much larger mixtures.
These findings motivate a growing body of work on principled data selection for instruction tuning.
Our method tackles this problem and enables data-efficient instruction tuning with a small fraction of the training pool.

\noindent\textbf{Data selection for instruction tuning.}
Existing methods~\citep{wu2024best,ivison2025large,liu2024tsds,cheng2026task,nayak2026critical,li2024one} for selecting instruction-tuning data can be broadly grouped into quality-based filtering and influence-based scoring~\citep{jiang2026importance}.
On the quality side, AlpaGasus~\citep{chen2023alpagasus} uses an LLM judge to filter low-quality examples, MoDS~\citep{du2023mods} introduces a three-stage pipeline based on quality, coverage, and necessity, IFD~\citep{li2024quantity} measures the gap between a model's expected and intrinsic responses to identify high-value examples, and DEITA~\citep{liu2023makes} scores data along complexity, quality, and diversity dimensions.
These methods rely on heuristic proxies and do not condition on a specific target capability.
On the influence side, influence functions~\citep{koh2017understanding} estimate the effect of removing a training point on model predictions via an inverse-Hessian--vector product.
TracIn~\citep{pruthi2020estimating} traces gradient inner products across training checkpoints, while TRAK~\citep{park2023trak} uses random projections for efficient attribution.
LESS~\citep{xia2024less} adapts this line of work to instruction tuning by combining LoRA-compressed gradients with Adam-adjusted influence scoring, enabling targeted selection where 5\% of the data suffices to match full-dataset performance.
In the multimodal setting, TIVE~\citep{liu2025less} extends gradient-based influence to visual instruction tuning.
However, all these methods produce point-wise scores that rank examples independently, ignoring inter-example interactions.
Our method builds on a submodular objective that explicitly models pairwise redundancy, while further closing coverage gaps via gradient-guided synthesis.

\noindent\textbf{Coreset selection and submodular optimization.}
Coreset selection aims to identify a representative subset that preserves the learning performance of the full dataset.
Classical approaches include geometry-based coverage~\citep{chen2012super_cs_1,agarwal2020contextual_cs_2,sener2017active,sinha2020small_cs_4}, uncertainty and error based sampling~\citep{coleman2019selection_cs5,paul2021deep_cs6,toneva2018empirical_cs7}, gradient-based methods~\citep{mirzasoleiman2020coresetscs_10,killamsetty2021grad_cs11} and decision-boundary-based methods~\citep{ducoffe2018adversarial_cs8,margatina2021active_cs9}.
Submodular functions provide a natural framework for subset selection with diminishing returns~\citep{nemhauser1978analysis,krause2014submodular}.
Wei et al.~\citep{wei2015submodularity} show that Facility Location and related coverage functions can select informative training subsets for supervised learning, and subsequent work applies submodular objectives to active learning~\citep{ash2019deep}, curriculum learning~\citep{killamsetty2021glister}, and continual learning~\citep{killamsetty2021retrieve}.
Graph-cut objectives have been used for document summarization~\citep{lin2011class} and general data subset selection~\citep{iyer2021submodular}.
We bring graph cuts into the query-based selection setting, replacing input features with gradient projections and coupling the dispersion penalty with an influence-derived relevance term so that the selected subset is both diverse and aligned with the target task.

\section{Method}
\label{sec:method}

\subsection{Problem Formulation}
Let $\mathcal{D} = \{\mathbf{x}_1, \dots, \mathbf{x}_n\}$ be a large pool of candidate instruction-tuning examples, and let $\mathcal{V}$ be a small target validation set containing a handful of examples that embody a specific desired capability.
The goal is to select a subset $\mathcal{S} \subseteq \mathcal{D}$ with $|\mathcal{S}| = K$ such that a model fine-tuned on $\mathcal{S}$ achieves low loss on $\mathcal{V}$, and consequently strong performance on the test set.
Directly finding the optimal subset is NP-hard. Most existing methods~\citep{xia2024less,xie2023data} operate under a point-wise assumption: they score each candidate independently against the target and select the top-$k$ examples.
However, by ignoring inter-sample relationships, this paradigm inevitably suffers from diversity collapse. When a target requires specific reasoning skills, the algorithm redundantly selects highly homogeneous variations of the same problem, wasting the limited budget.

\begin{figure}[t]
    \centering
    \includegraphics[trim=0 140 0 0, clip,width=1\linewidth]{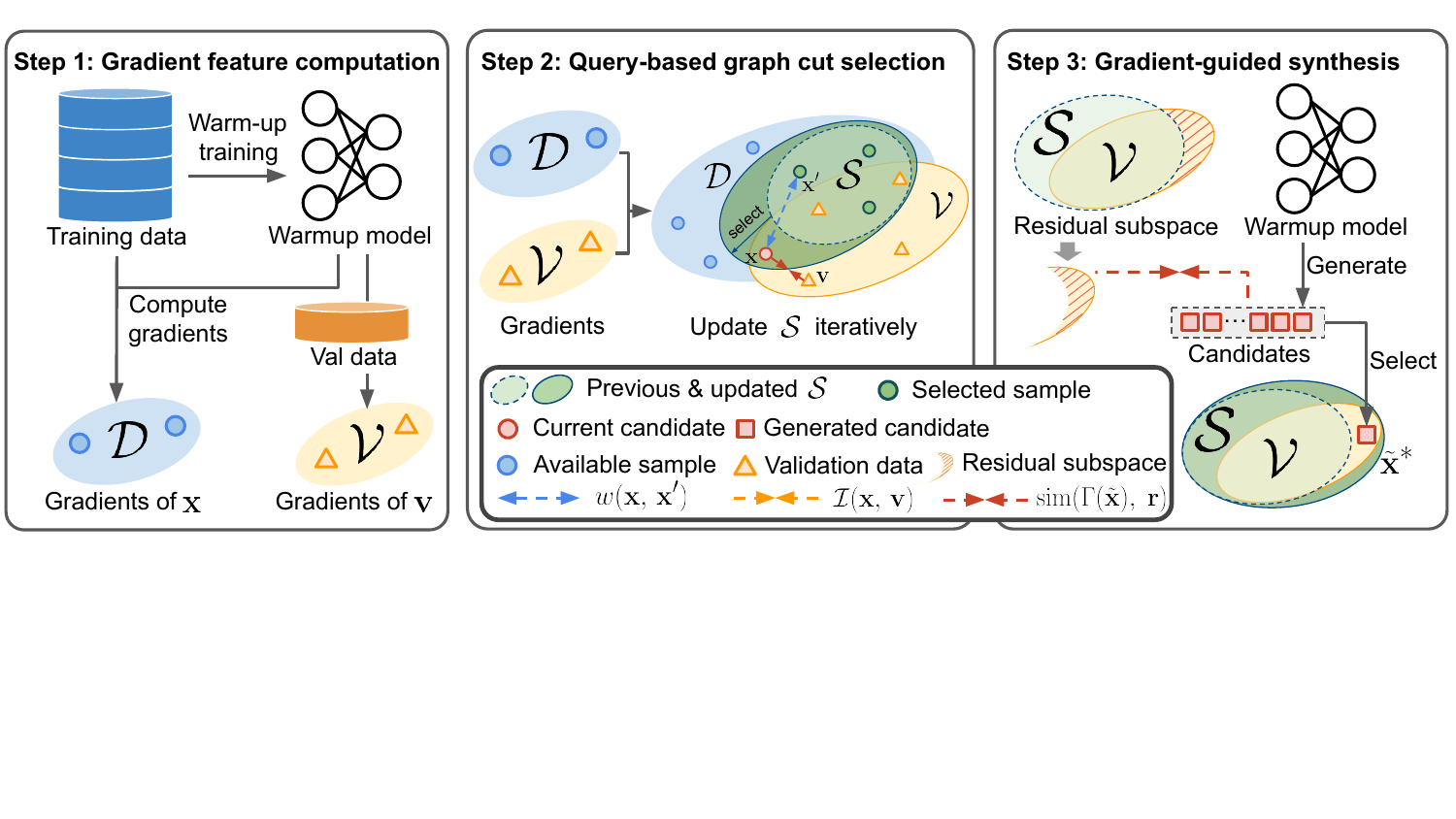}
    \vspace{-0.5cm}
    \caption{\textbf{Overview of the proposed QSGS framework.} Given a candidate pool $\mathcal{D}$ and a target validation set $\mathcal{V}$, QSGS constructs a high-quality coreset through three stages. \textbf{(1)~Gradient feature computation.} A warmup model is trained on a small random subset of the training data and used to compute per-sample gradient features for all examples in $\mathcal{D}$ and $\mathcal{V}$.    
    \textbf{(2)~Query-based graph cut selection.} For each candidate $\mathbf{x}$, we compute the similarity $w(\mathbf{x},\mathbf{x}')$ to already-selected samples, and the influence score $\mathcal{I}(\mathbf{x},\mathbf{v})$ measuring its relevance to the validation set. The submodular graph cut objective maximizes $\mathcal{I}(\mathbf{x},\mathbf{v})$ while penalizing $w(\mathbf{x},\mathbf{x}')$, jointly promoting candidate--target relevance and minimizing intra-subset redundancy to yield a diverse yet task-aligned coreset. 
    \textbf{(3)~Gradient-guided synthesis.} 
    To fill coverage gaps, we compute the gradient residual to find uncovered directions, then use the warmup model to generate synthetic candidates and select to the coreset.
    }
    \label{fig:structure}
    \vspace{-0.65cm}
\end{figure}

Targeting this problem, we propose QSGS framework as illustrated in Figure~\ref{fig:structure}.
The process for coreset construction can be summarized as follows:
(1)~\textbf{Gradient feature computation.} We compute low-dimensional gradient features for each data sample in $\mathcal{D}$ and derive influence scores that measure each candidate’s contribution to the target tasks.
(2)~\textbf{Query-based graph cut selection.} 
We formulate a submodular graph cut objective that jointly maximizes relevance to $\mathcal{V}$ and minimizes intra-subset redundancy in gradient space.
(3)~\textbf{Gradient-guided synthesis.} We identify residual subspace gaps
and generate targeted synthetic examples to fill them.

\subsection{Gradient Feature Computation}
\label{subsec:gradient}

We leverage gradient-based influence to measure each training sample’s contribution to the target tasks.
To make this tractable for large-scale datasets, we employ LoRA~\citep{hu2022lora} and random projections~\citep{johnson1984extensions} to reduce gradient dimensionality, and assign bigger weights to early-stage checkpoints before convergence to obtain more informative gradient features.
Crucially, we retain the pairwise influence score matrix rather than collapsing it into a scalar per sample, and construct compact gradient for efficient similarity computation in the subsequent graph cut stage.

Following~\citep{pruthi2020estimating}, we use a first-order approximation of training dynamics to estimate the influence of a training data point on held-out data.
For a warm-up model, the influence of a training example $\mathbf{x}$ on a validation example $\mathbf{v}$ is defined as:
\begin{equation}
  \mathcal{I}(\mathbf{x},\, \mathbf{v})
  \;=\;
  \sum_{i=1}^{T} \bar\eta_i \;
  \frac{\langle\, \Gamma(\mathbf{x};\, \theta_i),\;
    \Gamma(\mathbf{v};\, \theta_i) \,\rangle}
       {\|\Gamma(\mathbf{x};\, \theta_i)\|\, \|\Gamma(\mathbf{v};\, \theta_i)\|},
  \label{eq:influence}
\end{equation}
where $T$ is the number of warmup training epochs, $\bar\eta_i$ is the mean learning rate in epoch $i$, $\theta_i$ is the model checkpoint after epoch $i$, and $\Gamma(\cdot;\theta_i)$ denotes the projected Adam gradient update at checkpoint $\theta_i$.
Computing these scores requires model checkpoints along the training trajectory; we circumvent the circular dependency between data selection and training by performing a short warmup run on a randomly selected $\mathcal{D}_\text{warmup} \subset \mathcal{D}$.
Unlike prior work~\citep{xia2024less,zhang2025tagcos} that aggregates the per-validation-example scores into a single scalar per training sample,
we retain the full influence score matrix $\mathbf{S} \in \mathbb{R}^{n \times |\mathcal{V}|}$ with entries $S_{ij} = \mathcal{I}(\mathbf{x}_i, \mathbf{v}_j)$.
This 2D representation preserves the fine-grained structure of each candidate’s contribution to every validation example, which is essential for the relevance term in our graph cut objective (Section~\ref{subsec:graphcut}).

We perform LoRA fine-tuning~\citep{hu2022lora} on $\mathcal{D}_\text{warmup}$ for $T$ epochs, checkpointing the model after each epoch to store $\{\theta_i\}_{i=1}^T$ along with their Adam optimizer states.
LoRA freezes the pre-trained weights and adds a low-rank adaptor, reducing the number of trainable parameters to less than 2\% of the full model.
Each gradient is projected into a $d$-dimensional space via a fixed random matrix~\citep{johnson1984extensions,park2023trak}, forming a compact gradient datastore from which the influence score matrix $\mathbf{S}$ is computed.
Beyond the influence score matrix, the graph cut objective requires pairwise similarity between training examples.
We construct a gradient embedding for each training example by concatenating the checkpoint-weighted gradient projections:
\begin{equation}
  \mathbf{e}(\mathbf{x})
  \;=\;
  \Big[\,
    \sqrt{\bar\eta_1}\,\frac{\Gamma(x;\theta_1)}
{\|\Gamma(x;\theta_1)\|}
    \;\Big\|\;
    \cdots
    \;\Big\|\;
    \sqrt{\bar\eta_T}\,\frac{\Gamma(x;\theta_T)}
{\|\Gamma(x;\theta_T)\|}
  \,\Big]
  \;\in\; \mathbb{R}^{Td},
  \label{eq:embedding}
\end{equation}
where $\|$ denotes concatenation.
The $\sqrt{\bar\eta_i}$ weighting ensures that the inner product between two embeddings recovers the weighted cosine similarity used in the graph cut kernel (Eq.~\ref{eq:kernel}).
Since the concatenated dimension $Td$ can be large, we apply a Johnson--Lindenstrauss random projection~\citep{johnson1984extensions} to reduce each embedding to $d'$ dimensions while preserving pairwise distances with high probability.
The resulting embeddings $\tilde{\mathbf{e}}(\mathbf{x}) \in \mathbb{R}^{d'}$ are L2-normalized so that their inner product equals cosine similarity, enabling efficient similarity computation in the graph cut stage.

We propose assigning bigger weights to checkpoints sampled before convergence to compute gradient features.
This is motivated by the observation that gradient norms during the warmup phase are significantly larger than those at convergence, and therefore encapsulate more discriminative information about how each sample affects the model’s updates.

\subsection{Query-based Graph Cut Selection}
\label{subsec:graphcut}

To simultaneously combat diversity collapse and early saturation of coverage-based alternatives, we build on the classical submodular graph cut framework~\citep{iyer2021submodular} and adapt it to the query-driven data selection setting.
The standard graph cut objective selects a subset that minimizes intra-subset similarity, promoting diversity but without any notion of task relevance.
We introduce a query-aware relevance term derived from the influence score matrix that steers the selection toward the target validation set, turning the diversity-seeking cut into a joint relevance-diversity objective.

The relevance of a candidate $\mathbf{x}$ to the target set is the aggregate influence over all validation examples:
\begin{equation}
  \operatorname{Rel}(\mathbf{x},\, \mathcal{V})
  \;=\;
  \sum_{\mathbf{v} \in \mathcal{V}} \mathcal{I}(\mathbf{x},\, \mathbf{v}),
  \label{eq:relevance}
\end{equation}
measuring the total contribution of $\mathbf{x}$ to the entire target set.
The pairwise similarity between two candidates is defined by the weighted cosine similarity of their gradient projections across checkpoints:
\begin{equation}
    w(\mathbf{x},\, \mathbf{x}')
  \;=\;
  \sum_{i=1}^{T} \bar\eta_i \;
  \left(1+
  \frac{\langle \Gamma(\mathbf{x};\theta_i),\, \Gamma(\mathbf{x}';\theta_i) \rangle}
       {\|\Gamma(\mathbf{x};\theta_i)\|\, \|\Gamma(\mathbf{x}';\theta_i)\|},
  \label{eq:kernel}
    \right).
\end{equation}
The formula above ensure that $w(\mathbf{x}, \mathbf{x}') \geq 0$ for all pairs, which quantifies redundancy in gradient space: a high $w(\mathbf{x}, \mathbf{x}')$ indicates that the two examples induce similar parameter updates.

We seek a coreset $\mathcal{S} \subset \mathcal{D}$ of size $K$ that maintains high relevance to $\mathcal{V}$ while exhibiting low internal similarity.
We formalize this via a target-aware submodular graph cut:
\begin{equation}
  f(\mathcal{S})
  \;=\;
   \sum_{\mathbf{x} \in \mathcal{S}}\;
    \operatorname{Rel}(\mathbf{x},\, \mathcal{V})
  \;-\;
  \lambda \sum_{\substack{\mathbf{x}, \mathbf{x}' \in \mathcal{S} \\ \mathbf{x} \neq \mathbf{x}'}}
    w(\mathbf{x},\, \mathbf{x}'),
  \label{eq:objective}
\end{equation}
where the first term maximizes the cumulative influence of $\mathcal{S}$ on the target task, while the second term acts as a dispersion penalty that minimizes intra-subset redundancy in gradient space.
The hyperparameter $\lambda > 0$ balances relevance and diversity. Two terms capture complementary properties of a selected subset. The relevance term favors examples aligned with the target capability, while the pairwise term discourages repeatedly selecting examples that induce similar gradient updates. 
Under a tight cardinality constraint, each redundant example consumes a larger fraction of the available selection capacity and leaves fewer slots for complementary training signals. 
Furthermore, unlike coverage functions that saturate when the sparse set $\mathcal{V}$ is fully covered, our dispersion penalty operates on the internal structure of $\mathcal{S}$ itself and continues to discriminate among candidates regardless of $|\mathcal{V}|$.

Because $w(\mathbf{x},\, \mathbf{x}')\geq0$, the pairwise sum $\sum_{\mathbf{x},\mathbf{x}'\in S,\mathbf{x}\neq \mathbf{x}'}w(\mathbf{x},\, \mathbf{x}')$ is supermodular. Its negation is therefore submodular. Since the relevance term is modular, their sum $f(\mathcal{S})$ is submodular. The objective is generally nonmonotone because adding a highly redundant example can decrease its value.
The resulting objective is a nonmonotone submodular maximization problem under a cardinality constraint. Randomized algorithms~\citep{buchbinder2014submodular} provide constant factor approximation guarantees for this class of problems. In practice, we adopt the deterministic greedy procedure in Algorithm 1 because it is simple, efficient, and empirically stable for our selection setting. We therefore do not claim that the theoretical $1/e$ approximation guarantee applies to Algorithm 1.



\begin{wrapfigure}{l}{0.42\textwidth} 
\vspace{-0.1cm}
\begin{algorithm}[H] 
\small
\SetAlgoLined
\SetKwInOut{Input}{Input}
\SetKwInOut{Output}{Output}
\caption{Query-based graph cut}
\label{alg:subless}
\Input{
  Candidate pool $\mathcal{D}$;
  validation set $\mathcal{V} = \{\mathcal{V}_1,\dots,\mathcal{V}_m\}$;
  influence score matrix $\mathbf{S} \in \mathbb{R}^{n \times |\mathcal{V}|}$ with $S_{ij} = \mathcal{I}(\mathbf{x}_i, \mathbf{v}_j)$ \quad (Eq.~\ref{eq:influence});
  gradient embeddings $\tilde{\mathbf{e}}(\mathbf{x})$ for all $\mathbf{x} \in \mathcal{D}$ \quad (Eq.~\ref{eq:embedding});
  budget $K$
}
\Output{Coreset $\mathcal{S}$}

\For{each $\mathbf{x} \in \mathcal{D}$}{
  $\operatorname{Rel}(\mathbf{x}) \leftarrow \sum_{\mathbf{v} \in \mathcal{V}} \mathcal{I}(\mathbf{x}, \mathbf{v})$ \quad (Eq.~\ref{eq:relevance})\;
}
$\mathcal{S} \leftarrow \emptyset$\;
\While{$|\mathcal{S}| < K$}{
  $\mathbf{x}^* \leftarrow \operatorname{argmax}_{\mathbf{x} \in \mathcal{D} \setminus \mathcal{S}}\; \Delta(\mathbf{x} \mid \mathcal{S})$ \quad (Eq.~\ref{eq:marginal})\;
  $\mathcal{S} \leftarrow \mathcal{S} \cup \{\mathbf{x}^*\}$\;
}
\Return{$\mathcal{S}$}
\end{algorithm}
\vspace{-0.8cm}
\end{wrapfigure}
In practice, we use the deterministic greedy algorithm, as shown in Algorithm~\ref{alg:subless}.
At each iteration, we add the element with the largest marginal gain:
\begin{equation}
  \Delta(\mathbf{x}^* \mid \mathcal{S})
  \;=\;
  \operatorname{Rel}(\mathbf{x}^*,\, \mathcal{V})
  \;-\;
  2 \lambda\sum_{\mathbf{x}' \in \mathcal{S}} w(\mathbf{x}^*,\, \mathbf{x}').
  \label{eq:marginal}
\end{equation}
Intuitively, $\mathbf{x}^*$ is selected when its relevance to the target task outweighs its total similarity to examples already chosen, automatically balancing exploitation against exploration.
This criterion also makes the selection order interpretable, since each chosen example must provide additional target aligned information beyond what is already covered by $\mathcal{S}$.
To scale to large candidate pools, we approximate $w(\mathbf{x}, \mathbf{x}')$ via the compact gradient embeddings from Section~\ref{subsec:gradient}: 
$w(\mathbf{x}, \mathbf{x}') \approx \sum_{i=1}^{T}\bar{\eta}_i\left(1+\langle \hat{\mathbf{e}}(\mathbf{x}),\, \hat{\mathbf{e}}(\mathbf{x}') \rangle\right)$, where $\hat{\mathbf{e}}$ is the L2-normalized randomly projected embedding from Eq.~\eqref{eq:embedding}.
The additive constant contributes the same value to the marginal gain of every remaining candidate at a fixed greedy iteration and therefore does not affect the selection ordering. This avoids recomputing over the raw gradients across all checkpoint gradients. 

\subsection{Gradient-Guided Synthesis}
\label{subsec:synthesis}

Even with diversity-aware selection, the training pool $\mathcal{D}$ may not contain examples that cover every gradient direction required by the target task.
At extreme budgets, this subspace gap is especially acute: the selected subset $\mathcal{S}$ spans a low-dimensional subspace of gradient space, and some target capabilities may lie entirely outside it.
We address this with a post-selection synthesis step that uses the gradient geometry of $\mathcal{S}$ to identify what is missing and generate targeted synthetic examples.

\noindent\textbf{Residual subspace identification.}
After selecting $\mathcal{S}$ via Algorithm~\ref{alg:subless}, we measure how well the selected subset covers the target validation set in gradient space.
For target validation set $\mathcal{V}$, we compute the gradient residual:
\begin{equation}
  \mathbf{r} \;=\; \bar{g}(\mathcal{V}) \;-\; \operatorname{proj}_{\operatorname{span}(\Gamma(\mathcal{S}))}\, \bar{g}(\mathcal{V}),
  \label{eq:residual}
\end{equation}
where $\bar{g}(\mathcal{V}) = \frac{1}{|\mathcal{V}|}\sum_{v \in \mathcal{V}} \Gamma(v)$ is the mean gradient of the validation set, which is averaged over checkpoints for brevity, and the projection is onto the subspace spanned by $\{\Gamma(\mathbf{x}) : \mathbf{x} \in \mathcal{S}\}$.
The residual norm $\|\mathbf{r}\|$ quantifies the uncovered component of validation set: a large residual indicates that no selected example provides a gradient update aligned with this target direction.

\noindent\textbf{Targeted synthesis.}
We leverage the warm-up model $\theta$ to generate candidate synthetic samples with the validation examples from $\mathcal{V}$.
The prompt instructs the model to produce diverse instruction--response pairs that exercise the same capability from different surface contexts.
We then compute the gradient features of each synthetic candidate $\tilde{\mathbf{x}}$ using the same LoRA checkpoints and random projection as the original datastore, and select the candidates whose gradient direction best aligns with the residual:
\begin{equation}
  \tilde{\mathbf{x}}^* \;=\; \operatorname{argmax}_{\tilde{\mathbf{x}}}\;
  \frac{\langle\, \Gamma(\tilde{\mathbf{x}}),\; \mathbf{r} \,\rangle}
       {\|\Gamma(\tilde{\mathbf{x}})\|\, \|\mathbf{r}\|}.
  \label{eq:synth_select}
\end{equation}
Selected synthetic examples are added to $\mathcal{S}$, the residuals are recomputed, and the process repeats until a synthesis budget $K_\text{syn}$ is exhausted or all residual norms fall below a threshold $\epsilon$.

Na\"ive synthetic data generation produces examples that are semantically related to the target task but may be redundant with data already in $\mathcal{S}$.
By selecting synthetic candidates based on their alignment with the residual gradient direction, we ensure that each synthetic example provides a training signal that is both target-relevant and complementary to the existing subset.
This couples the generation process tightly with the graph cut framework: the residual in Eq.~\eqref{eq:residual} is precisely the component of the target gradient that the dispersion-optimized subset failed to capture.
\section{Experiments}
\label{sec:experiments}
\vspace{-0.2cm}

We first describe the experimental setup in Section~\ref{subsec:setup}, then present main results in Section~\ref{subsec:main_results}. Section~\ref{subsec:ablation} provides ablation studies isolating the contribution of each component, and offers further analysis including comparisons of submodular objectives.

\vspace{-0.25cm}
\subsection{Experimental Setup}
\label{subsec:setup}
\vspace{-0.15cm}

\noindent\textbf{Dataset and metrics.}
Following~\citep{xia2024less}, we use a heterogeneous pool of ${\sim}270$K instruction-tuning examples drawn from four sources: Flan~v2~\citep{longpre2023flan}, Chain-of-Thought~\citep{wei2022chain}, Dolly~\citep{DatabricksBlog2023DollyV2}, and Open Assistant~\citep{kopf2023openassistant}.
This pool spans a wide range of formats and reasoning types and contains no obvious in-domain data for the target evaluation tasks, making it a realistic and challenging selection environment.
We evaluate on three benchmarks that together cover diverse capability dimensions:
MMLU~\citep{hendrycks2020measuring} comprises 57 multiple-choice tasks spanning science, humanities, and professional knowledge;
BBH~\citep{suzgun2023challenging} contains 27 hard reasoning tasks selected from BIG-Bench;
TydiQA~\citep{tydiqa} is a multilingual extractive QA dataset covering 9 typologically diverse languages.
For each benchmark, the few-shot examples provided per subtask serve as the validation set $\mathcal{V}$ for data selection and as in-context demonstrations during evaluation. 

\noindent\textbf{Implementation details.}
Warmup training uses a randomly selected 5\% of $\mathcal{D}$ for $T=4$ epochs, producing 4 LoRA checkpoints.
Gradients are projected to $d=8192$ dimensions.
Unless otherwise noted, we report results at a 0.5\% data budget ($K \approx 1{,}350$ examples), the extreme-budget regime that is our primary focus.
We additionally evaluate at 1\%, 2\%, and 5\% budgets for the scaling study.
For gradient-guided synthesis, we generate $K_\text{syn} = 100$ candidate examples per under-covered subtask and select up to 10 synthetic examples total. For LLama 3 training, following~\citep{wang2023far, ivison2023camels, zhang2025tagcos}, we extend the training set to a combination of 17 datasets. All experiments are conducted on NVIDIA RTX A6000 GPUs.
We evaluate QSGS across three base models: Llama-2-7B~\citep{touvron2023llama}, Llama-3-8B~\citep{grattafiori2024llama}, Mistral-7B~\citep{jiang2023mistral}.
All warmup and final fine-tuning are performed with LoRA~\citep{hu2022lora}. All ablations are conducted on Llama-2-7B / MMLU at the 0.5\% budget.

\vspace{-0.25cm}
\subsection{Main Results}
\label{subsec:main_results}
\vspace{-0.15cm}

\begin{table}[t]
\caption{%
  Coreset selection performance at the 0.5\% data budget across three models and three benchmarks.
  Full denotes training on all ${\sim}270$K examples;
  Rand.\ denotes uniform 0.5\% random sampling;
  LESS~\citep{xia2024less} denotes top-$K$ gradient influence selection.
  \textbf{Bold} indicates the best result among selected subsets.
  Standard deviations over three seeds are reported in subscript.
  QSGS consistently outperforms all baselines across every model--benchmark combination.
}
\vspace{-0.3cm}
\label{tab:main_result}
\resizebox{\textwidth}{!}{%
\setlength{\tabcolsep}{3.5pt}
\begin{tabular}{lcccc|cccc|cccc}
\toprule
 & \multicolumn{4}{c|}{\textbf{MMLU}} & \multicolumn{4}{c|}{\textbf{TydiQA}} & \multicolumn{4}{c}{\textbf{BBH}} \\
\cmidrule(lr){2-5}\cmidrule(lr){6-9}\cmidrule(lr){10-13}
\textbf{Model}
  & \textbf{Full} & \textbf{Rand.} & \textbf{LESS} & \textbf{QSGS}
  & \textbf{Full} & \textbf{Rand.} & \textbf{LESS} & \textbf{QSGS}
  & \textbf{Full} & \textbf{Rand.} & \textbf{LESS} & \textbf{QSGS} \\
\textbf{Data \%}
  & \small(100\%) & \small(0.5\%) & \small(0.5\%) & \small(0.5\%)
  & \small(100\%) & \small(0.5\%) & \small(0.5\%) & \small(0.5\%)
  & \small(100\%) & \small(0.5\%) & \small(0.5\%) & \small(0.5\%) \\
\cmidrule(r){1-1}
\cmidrule(lr){2-2}\cmidrule(lr){3-5}
\cmidrule(lr){6-6}\cmidrule(lr){7-9}
\cmidrule(lr){10-10}\cmidrule(lr){11-13}
\textbf{Llama-2-7B}
  & \textit{51.6}
  & 45.0\std{1.3} & 45.2\std{0.5} & \textbf{47.9}\std{0.4}
  & \textit{54.0}
  & 48.4\std{0.9} & {50.2}\std{0.2} & {\textbf{52.3}}\std{0.2}
  & \textit{43.2}
  & 35.7\std{0.6} & 38.3\std{0.5} & \textbf{40.1}\std{0.6} \\
\textbf{Mistral-7B}
  & \textit{60.4}
  & 57.0\std{0.5} & {58.8}\std{0.7} & {\textbf{60.4}}\std{0.7}
  & \textit{57.7}
  & 53.9\std{0.7} & {55.3}\std{0.5} & {\textbf{57.5}}\std{0.3}
  & \textit{53.0}
  & 51.5\std{0.6} & {52.4}\std{0.2} & {\textbf{53.2}}\std{0.8} \\
\textbf{Llama-3-8B}
  & \textit{70.4}
  & 67.0\std{0.4} & 67.5\std{0.8} & \textbf{68.3}\std{0.6}
  & \textit{70.7}
  & 67.1\std{0.6} & 67.7\std{0.2} & \textbf{68.7}\std{0.3}
  & \textit{65.6}
  & 62.3\std{1.3} & 62.9\std{1.1} & \textbf{63.9}\std{0.8} \\
\bottomrule
\end{tabular}}

\vspace{0.2cm}
\begin{minipage}[t]{0.635\columnwidth}
\caption{%
  Full baseline comparison at 0.5\% data budget on Llama-2-7B. Our method outperforms existing methods.
}
\label{tab:baselines}
\vspace{-0.3cm}
\centering
\resizebox{\textwidth}{!}{%
\setlength{\tabcolsep}{3pt}
\begin{tabular}{lccccccc}
\toprule
 & \textbf{Rand.} & \textbf{BM25} & \textbf{DSIR} & \textbf{RDS}
 & \textbf{LESS} & \textbf{TAGCOS} & \textbf{QSGS} \\
\midrule
\textbf{MMLU}
  & 45.0\std{1.3} & 46.6\std{0.5} & 45.1\std{0.1} & 45.0\std{0.3}
  & 45.2\std{0.5} & 44.8\std{0.3} & \textbf{47.9}\std{0.4} \\
\textbf{TydiQA}
  & 48.4\std{0.9} & 49.7\std{0.7} & 47.5\std{0.5} & 48.6\std{0.1}
  & 50.2\std{0.2} & 48.1\std{0.3} & \textbf{52.3}\std{0.2} \\
\textbf{BBH}
  & 35.7\std{0.6} & 39.8\std{0.5} & 36.8\std{0.5} & 36.7\std{0.2}
  & 38.3\std{0.5} & 39.1\std{0.7} & \textbf{40.1}\std{0.6} \\
\bottomrule
\end{tabular}}
\end{minipage}%
\hfill
\begin{minipage}[t]{0.345\columnwidth}
\caption{%
  MMLU evaluation on under different data ratios.
}
\label{tab:data_ratio}
\vspace{-0.3cm}
\centering
\resizebox{\textwidth}{!}{
\setlength{\tabcolsep}{3pt}
\begin{tabular}{lccc}
\toprule
\textbf{Data Ratio} & \textbf{5\%} & \textbf{1\%} & \textbf{0.5\%} \\
\midrule
\textbf{Rand.}   & 46.5\std{0.4} & 45.9\std{0.6} & 45.0\std{1.3} \\
\textbf{LESS}    & 50.2\std{0.5} & 46.3\std{0.7} & 45.2\std{0.5} \\
\textbf{QSGS}  & \textbf{51.1}\std{0.3} & \textbf{48.5}\std{0.2} & \textbf{47.9}\std{0.4} \\
\bottomrule
\end{tabular}}
\end{minipage}
\vspace{-0.5cm}
\end{table}

Table~\ref{tab:main_result} presents results at the 0.5\% data budget ($K \approx 1{,}350$ examples).
QSGS improves over LESS by 1.6\% on average across all nine model--benchmark combinations, with gains in every individual setting.
The improvement is most pronounced on Llama-2-7B, where QSGS leads by 2.2\% on average and the gap between full-data and 0.5\% performance is also the widest.
This is consistent with our expectation: when the budget is tight relative to model capacity, top-$K$ selection is more prone to spending its limited budget on redundant high-scoring examples, and diversity-aware selection yields the greatest marginal benefit.
On the stronger Llama-3-8B, the absolute gain narrows to 0.9\%, suggesting that a more capable base model is more tolerant of suboptimal data composition.
Notably, on Mistral-7B, QSGS at 0.5\% matches or approaches full-data performance on all three benchmarks, closing the gap to within 0.2\% on each.
This suggests that for sufficiently strong base models, a small but well-diversified subset can recover nearly all the knowledge that 270K heterogeneous examples provide, as most of the full training pool is redundant.
Across all models, QSGS closes roughly half the gap between LESS and the full-data baseline.

We further compare QSGS against a broader set of non-gradient baselines on Llama-2-7B in Table~\ref{tab:baselines}. Lexical retrieval methods such as BM25~\citep{robertson2009probabilistic} and representation-based approaches including DSIR~\citep{xie2023data} and RDS~\citep{hanawa2020evaluation} all perform close to random selection, trailing QSGS by 1.3\% to 2.6\%.
These methods score each candidate independently against the target distribution and lack any mechanism to enforce diversity within the selected subset. As a result, they tend to retrieve clusters of near-duplicate examples that are individually relevant but jointly redundant, exhausting the limited budget without adequate coverage of the target gradient space.
In contrast, the graph cut objective in QSGS explicitly penalizes pairwise similarity among selected samples, ensuring that each additional example contributes complementary gradient information rather than duplicating directions already represented in the coreset.

Table~\ref{tab:data_ratio} traces the performance gap between QSGS and LESS on MMLU as the data budget decreases from 5\% to 0.5\%.
At 5\%, the two methods differ by only 0.9\%, as the budget is large enough for top-$K$ to incidentally achieve reasonable diversity.
However, the gap widens steadily to 2.2\% at 1\% and 2.7\% at 0.5\%.
The increasing empirical gap is consistent with the intuition that redundancy is more costly when only a small number of examples can be retained. With a larger budget, top-K selection can include redundant examples while still covering many useful gradient directions. Under a tight budget, each redundant selection directly displaces a potentially complementary example.
The submodular objective of QSGS directly addresses this by balancing relevance and coverage, allowing it to maintain broad gradient coverage even under tight budgets.
Quantitatively, LESS degrades by 5.0\% from the 5\% budget to the 0.5\% budget, while QSGS degrades by only 3.2\% over the same range.
Notably, QSGS at 0.5\% budget already surpasses LESS at 1\% budget.

\vspace{-0.25cm}
\subsection{Ablation Studies}
\label{subsec:ablation}
\vspace{-0.15cm}

\begin{figure*}[t]
  \centering
  \begin{subfigure}{0.24\linewidth}
    \centering
    \captionsetup{justification=centering}
    \includegraphics[trim=0 0 0 0,clip,width=\linewidth]{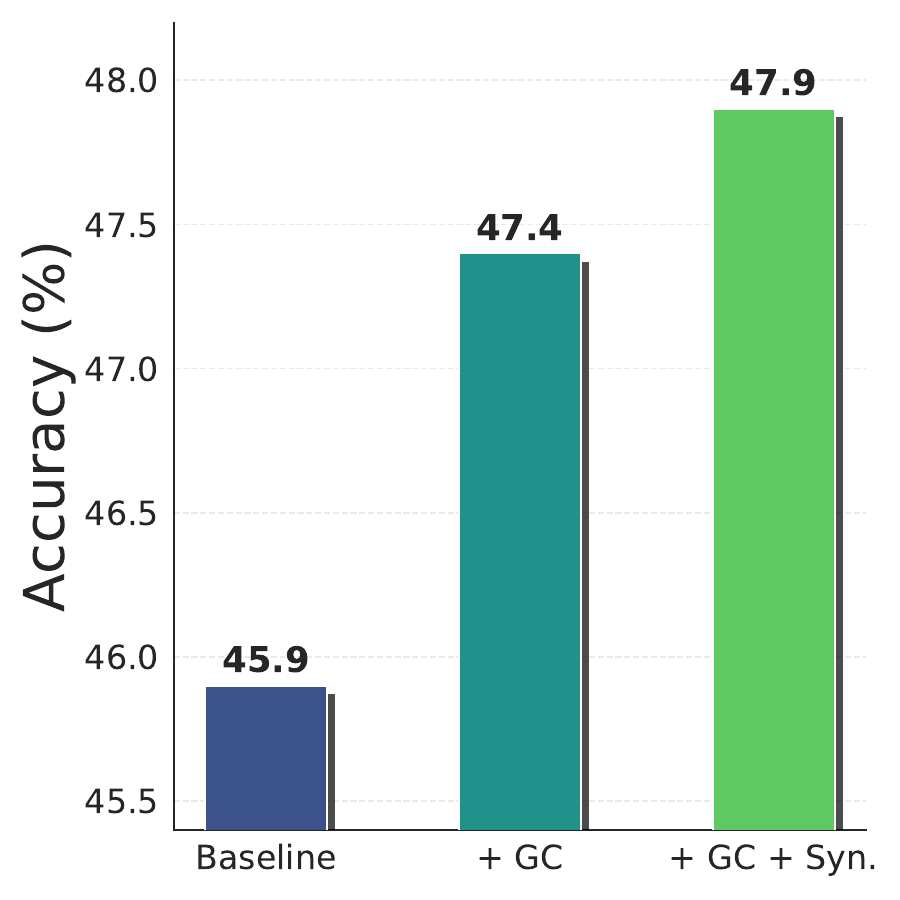}
    \vspace{-0.2in}
    \caption{Component analysis}
    \label{fig:abl_component}
  \end{subfigure}
  \hfill
  \begin{subfigure}{0.24\linewidth}
    \centering
    \captionsetup{justification=centering}
    \includegraphics[trim=0 0 0 0,clip,width=\linewidth]{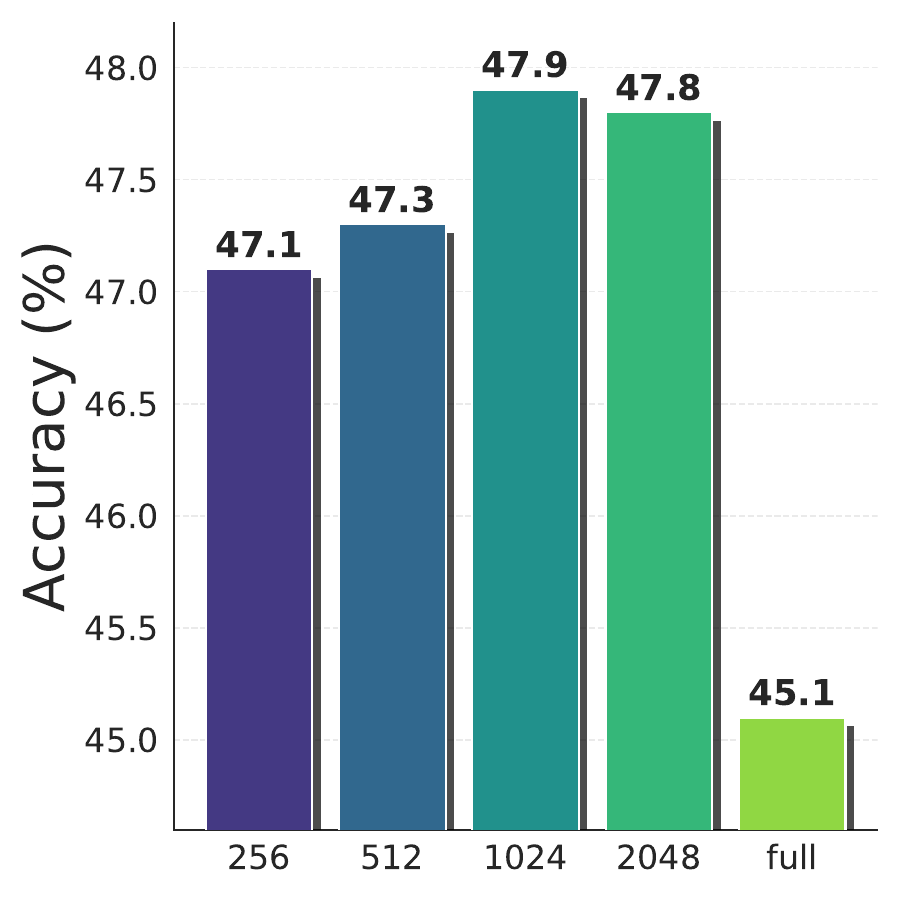}
    \vspace{-0.2in}
    \caption{Projection dimension $d$}
    \label{fig:abl_rp}
  \end{subfigure}
  \hfill
  \begin{subfigure}{0.24\linewidth}
    \centering
    \captionsetup{justification=centering}
    \includegraphics[trim=0 0 0 0,clip,width=\linewidth]{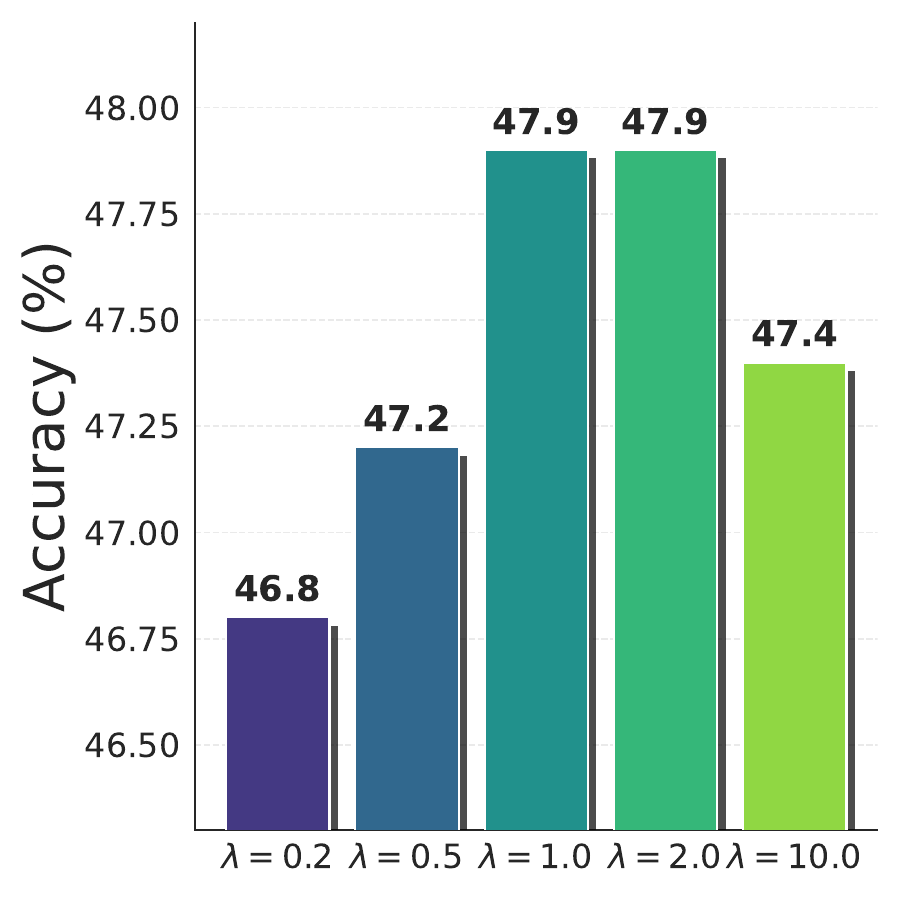}
    \vspace{-0.2in}
    \caption{Trade-off parameter $\lambda$}
    \label{fig:abl_lambda}
  \end{subfigure}
  \hfill
  \begin{subfigure}{0.24\linewidth}
    \centering
    \captionsetup{justification=centering}
    \includegraphics[trim=0 0 0 0,clip,width=\linewidth]{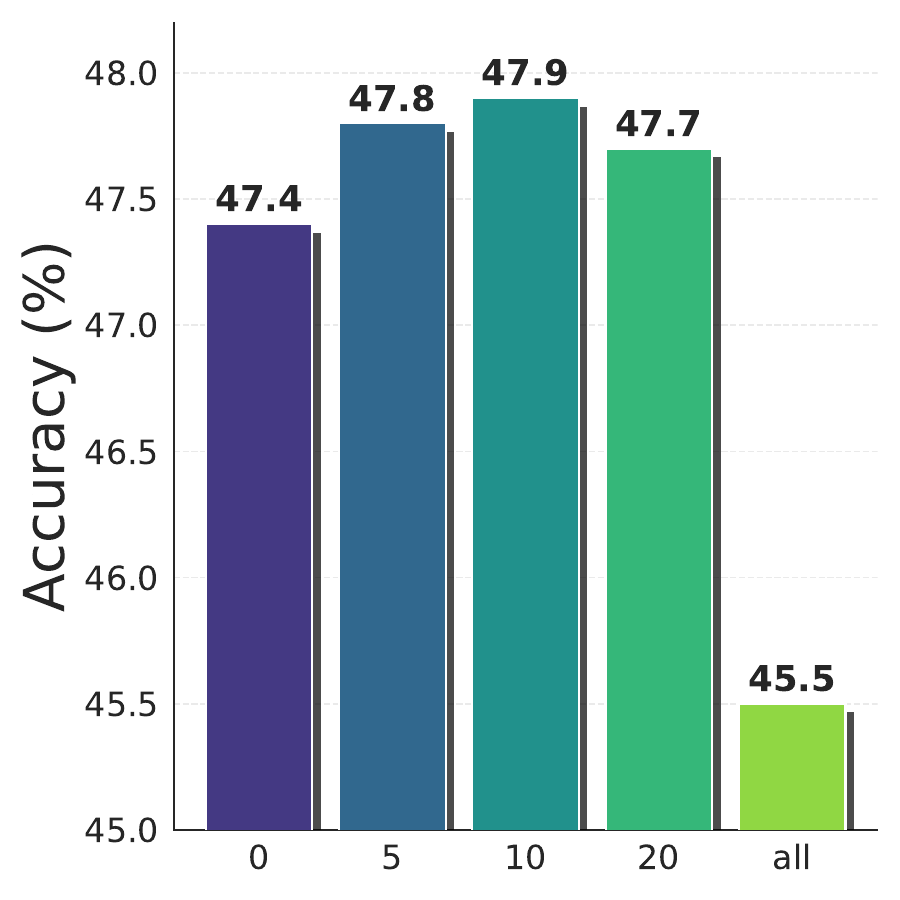}
    \vspace{-0.2in}
    \caption{Number of synthetics}
    \label{fig:abl_number}
  \end{subfigure}
  \vspace{-6pt}
  \caption{\textbf{Ablation studies} on Llama-2-7B / MMLU at the 0.5\% budget. \textbf{(a)} Incremental effect of each component. 
  \textbf{(b)} Effect of random projection dimension in the Graph Cut kernel. \textbf{(c)} Sensitivity to the relevance--diversity trade-off parameter $\lambda$. \textbf{(d)} Effect of synthetic samples added to coreset}
  \label{fig:ablation}
  \vspace{-0.7cm}
\end{figure*}

\noindent\textbf{Component analysis.}
To isolate the contribution of each proposed component, we incrementally add them to our baseline (Figure~\ref{fig:abl_component}). The baseline relies entirely on top-$K$ selection using computed gradient features, without any synthetic samples. We evaluate two extensions: ``+GC'', which replaces top-$K$ with our query-based Graph Cut, and ``+GC+Syn'', which represents the full method by adding gradient-guided sample synthesis.Replacing the baseline's top-$K$ selection with the Graph Cut objective accounts for the majority of the performance gain, increasing MMLU accuracy from 45.9 to 47.4. This confirms that diversity-aware selection is the primary driver of improvement. Building on this, gradient-guided synthesis provides a complementary boost by filling residual gaps in gradient coverage that the existing pool cannot address. This diminishing return from synthesis is entirely by design: because it specifically targets the subspace not spanned by pool-selected examples, its impact naturally shrinks as the quality of the initial pool selection improves.

\noindent\textbf{Random projection dimension.}
We vary the projection dimension $d'$ of the compact gradient embeddings $\tilde{\mathbf{e}}(\mathbf{x}) \in \mathbb{R}^{d'}$ used for pairwise similarity computation in the Graph Cut objective (Eq.~\ref{eq:embedding}).
Performance peaks at $d'{=}1024$ and plateaus at $d'{=}2048$ (Figure~\ref{fig:abl_rp}). For full-dimensional gradients, we use a chunked CPU--GPU transfer to accommodate the full vectors in memory during selection. Notably, operating on full-dimensional gradients degrades performance to 45.1, below random projection at every tested dimension.
We attribute this to distance concentration in high-dimensional spaces. As the gradient dimensionality grows, pairwise cosine similarities concentrate near zero, causing the redundancy penalty in the Graph Cut objective to vanish.
When the penalty term becomes negligible, Graph Cut effectively reduces to top-$K$ relevance ranking and loses its diversity-enforcing capability.
Random projection into a lower-dimensional subspace compresses the gradient representations so that semantically similar examples exhibit meaningfully non-zero similarity, thereby restoring the discriminative power of the redundancy term. Random projection reduces computational cost and acts as an implicit regularizer that preserves dominant gradient structure while suppressing the high-dimensional noise that would neutralize the diversity objective.

\noindent\textbf{Sensitivity to $\lambda$.}
Performance varies by only 1.1\% across a $50\times$ range of $\lambda$ (Figure~\ref{fig:abl_lambda}). The optimum lies at a small $\lambda$ where relevance remains the dominant selection criterion and the redundancy penalty acts only as a mild corrective. It is sufficient to suppress near-duplicate examples whose gradient directions are nearly identical, yet too weak to override relevance-based ranking.
This regime retains the highest-quality, task-aligned examples while eliminating pure redundancy, allowing the model to converge efficiently without wasting capacity on repeated training signal.
At the other extreme, $\lambda{=}10.0$ over-weights the diversity penalty, forcing the objective to sacrifice relevance for dissimilarity and thereby selecting uninformative examples.

\noindent\textbf{Number of synthesized candidates.}
We vary the number of synthetic samples added to the coreset after Graph Cut selection (Figure~\ref{fig:abl_number}).
Performance peaks at 10 samples and degrades sharply when all candidates are included, even falling below the no-synthesis baseline.
This is consistent with the residual-guided selection in Eq.~\eqref{eq:synth_select}: the first few synthetic examples align with the largest uncovered gradient directions and provide genuinely complementary signal; once these gaps are filled, additional candidates inevitably overlap with existing coreset members and dilute the diversity that Graph Cut optimization establishes.

\begin{wrapfigure}{r}{0.5\textwidth}
    \begin{subfigure}{0.48\linewidth}
    \centering
    \captionsetup{justification=centering}
    \vspace{-0.2in}
    \includegraphics[trim=0 0 0 0,clip,width=\linewidth]{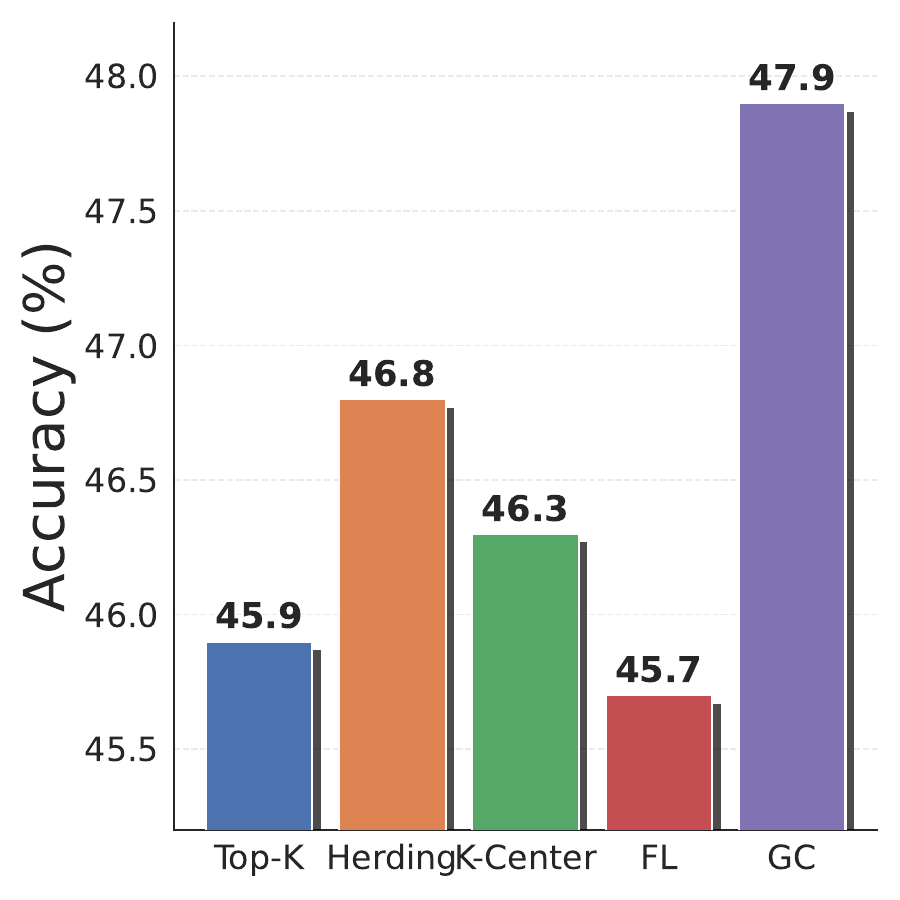}
    \vspace{-0.2in}
    \caption{Comparison of selection methods}
    \label{fig:anl_select}
  \end{subfigure}
  \hfill
  \begin{subfigure}{0.48\linewidth}
    \centering
    \captionsetup{justification=centering}
    \vspace{-0.2in}
    \includegraphics[trim=0 0 0 0,clip,width=\linewidth]{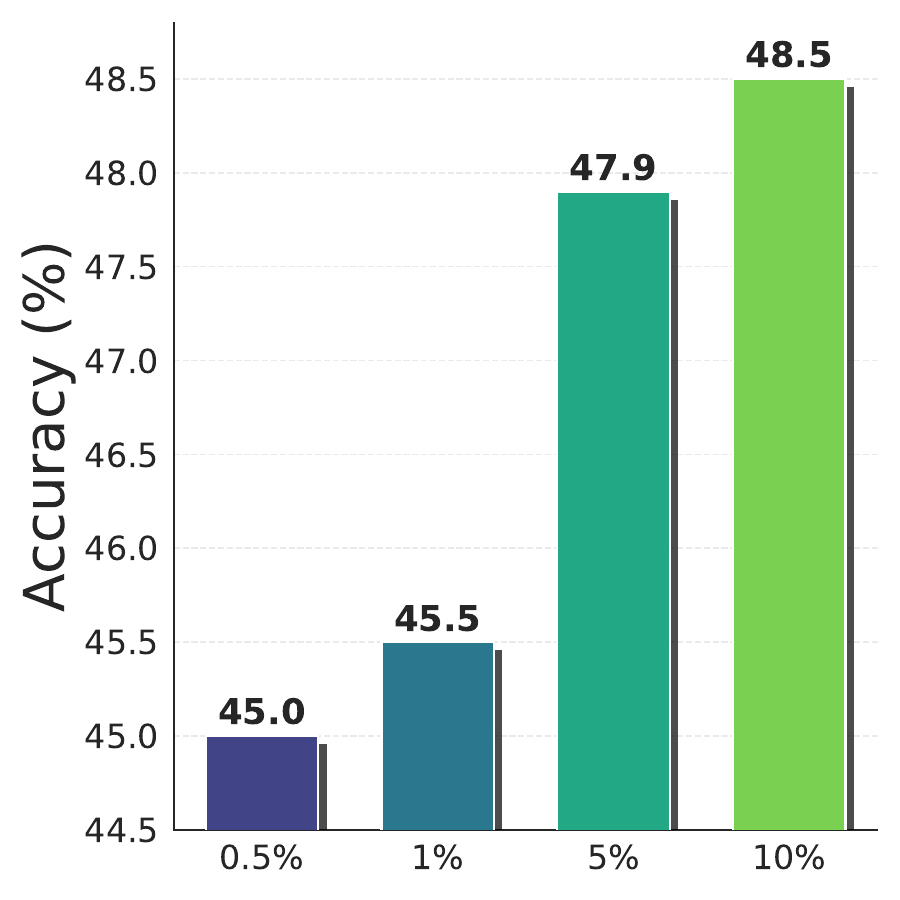}
    \vspace{-0.2in}
    \caption{Warmup data ratio influence}
    \label{fig:abl_warmup}
  \end{subfigure}
  \vspace{-0.3cm}
  \caption{Analysis on MMLU at the 0.5\% budget.}
  \vspace{-0.5cm}
\end{wrapfigure}

\noindent\textbf{Effect of selection strategy.}
We replace QSGS's Graph Cut (GC) objective with alternative submodular functions, namely Herding, $K$-Center, and Facility Location (FL), while keeping the same candidate buffer and gradient features, to isolate the role of the objective design. GC achieves the best performance, outperforming Herding and $K$-Center by 1.1\% and 1.6\%, respectively. Most notably, FL performs worse than even the top-$K$ baseline.
We trace this to the small size of the validation set: FL saturates coverage after selecting only ${\sim}$10 examples, at which point the marginal gain of every remaining candidate drops to near zero and subsequent selections are effectively random.
The root cause lies in FL's $\max$-based aggregation, which scores each candidate by its similarity to the single closest selected example. Once every validation point has a nearby representative, the marginal gain of additional candidates vanishes.
GC avoids this saturation by computing cumulative similarity between all selected samples and the validation set, so each new addition continues to yield positive marginal gains. 

\noindent\textbf{Warmup data ratio.}
Performance improves steadily as the warmup ratio increases from 0.5\% to 10\%, reflecting higher-quality gradient features (Figure~\ref{fig:abl_warmup}).
We adopt 5\% as the default: the gain from 5\% to 10\% is only 0.6\% while doubling the warmup cost, indicating diminishing returns beyond 5\%.
The sharp drop at 0.5\% warmup, which falls below even random selection, underscores that gradient-based selection fundamentally depends on the fidelity of the influence estimates; noisy gradients make even diversity-aware optimization ineffective.
\vspace{-0.4cm}
\section{Conclusion}
\vspace{-0.3cm}

We presented QSGS, a coreset construction framework for targeted instruction tuning that combines query-based submodular data selection with gradient-guided data synthesis.
The graph cut objective jointly maximizes target relevance and minimizes intra-subset redundancy. Empirically, its advantage over pointwise selection becomes increasingly pronounced under tighter data budgets, where redundant selections consume a larger fraction of the available training capacity.
A complementary synthesis step fills residual coverage gaps by generating examples aligned with uncovered target gradient directions.
Experiments across three benchmarks and three model families show that QSGS at a 0.5\% budget consistently outperforms all baselines, achieving full-data performance on Mistral-7B.

\bibliography{iclr2027_conference}
\bibliographystyle{iclr2027_conference}

\newpage
\appendix
\section{More Analysis}

\noindent\textbf{Computational overhead.}
We report the wall-clock cost of each pipeline stage on NVIDIA RTX A6000 GPUs in Figure 5a. Warmup LoRA training requires approximately 304 GPU hours, gradient feature construction approximately 48 GPU hours, data construction including graph cut selection and optional synthesis less than one GPU hour, and final fine-tuning on the selected coreset approximately 32 GPU hours. 
Importantly, the warmup training and gradient computation are one-time costs that can be fully amortized: once the gradient features are stored, the data construction stage can be repeated in under 1 hour to produce coresets for different target validation sets, different data budgets, or different submodular trade-off parameters, without re-running any GPU-intensive computation.
QSGS is therefore particularly suitable for settings in which multiple target-specific subsets are constructed from the same candidate pool.

\begin{wrapfigure}{r}{0.6\textwidth}
    \begin{subfigure}{0.48\linewidth}
    \centering
    \captionsetup{justification=centering}
    \vspace{-0.5in}
    \includegraphics[trim=0 0 0 0,clip,width=\linewidth]{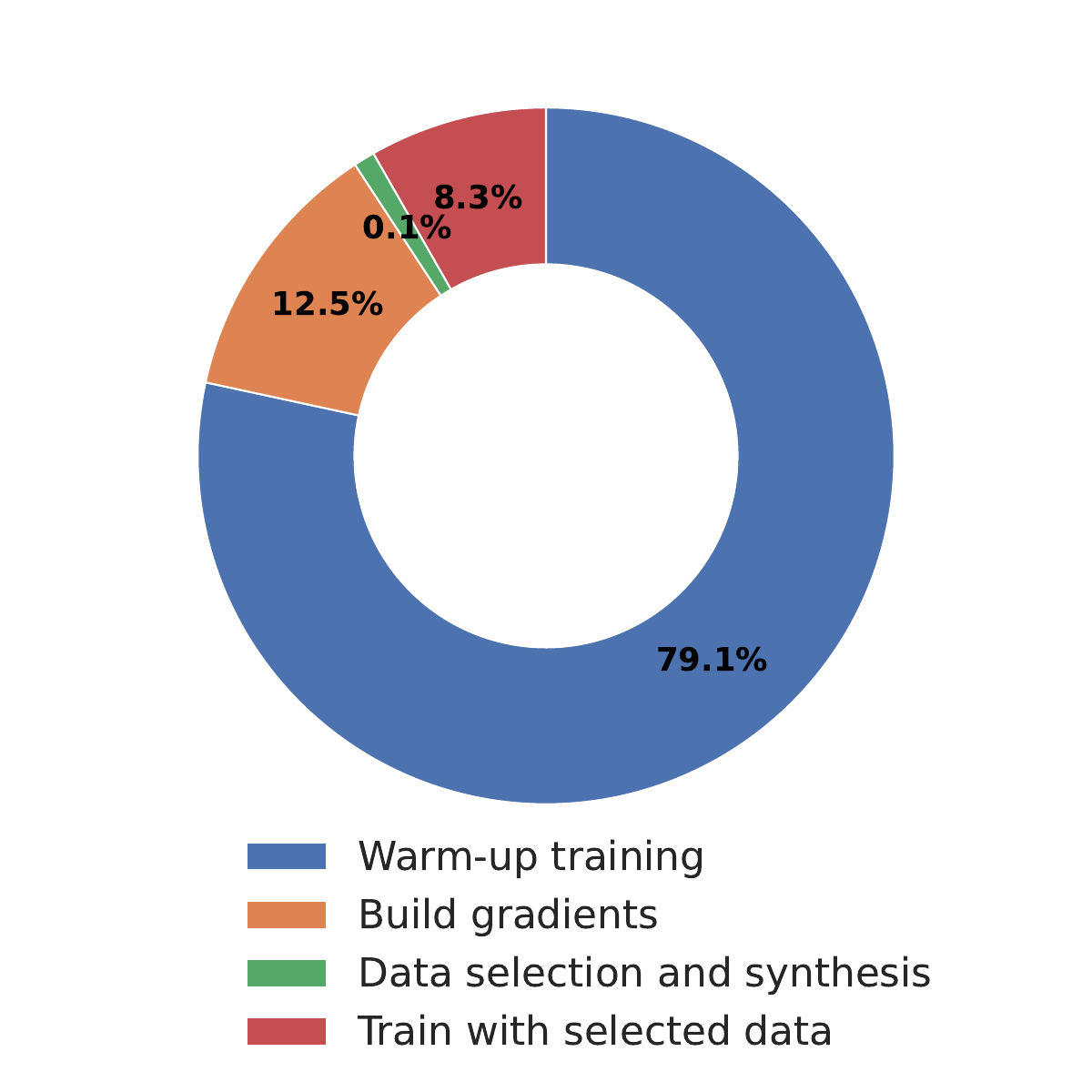}
    \vspace{-0.2in}
    \caption{Computational cost.}
    \label{fig:anl_compute}
  \end{subfigure}
  \hfill
  \begin{subfigure}{0.48\linewidth}
    \centering
    \captionsetup{justification=centering}
    \includegraphics[trim=0 40 310 0,clip,width=\linewidth]{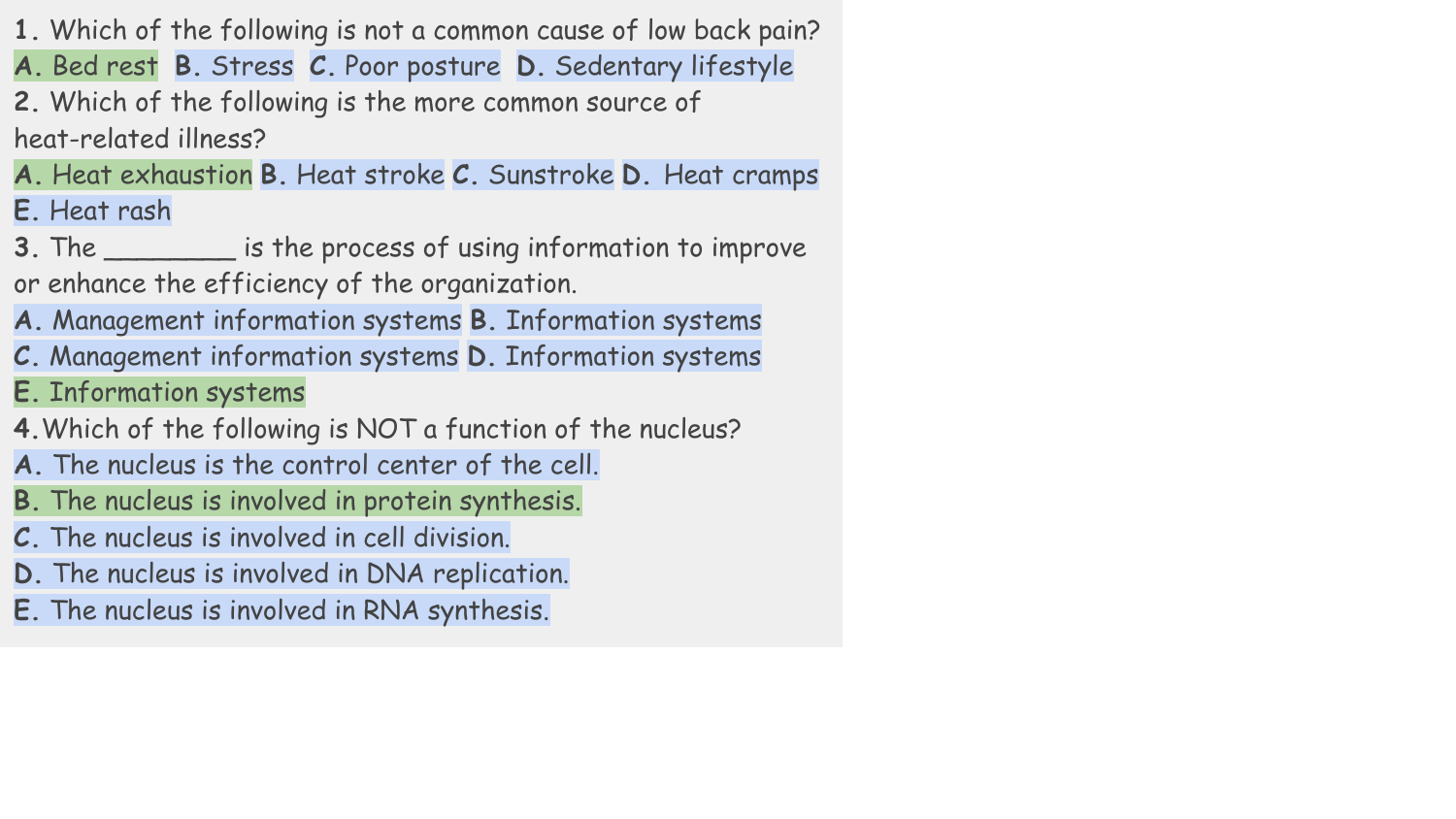}
    \vspace{-0.2in}
    \caption{Generated samples.}
    \label{fig:qualitative}
  \end{subfigure}
  \caption{Analysis on Llama-2-7B / MMLU at the 0.5\% budget. \textbf{(a)} Computational cost breakdown of each pipeline stage. \textbf{(b)} Representative synthetic examples generated for under-covered subtasks.}
  \vspace{-0.2cm}
\end{wrapfigure}
\noindent\textbf{Generated candidate examples.}
Figure~\ref{fig:qualitative} shows representative candidates produced for MMLU subtasks identified as under-covered by the residual analysis. The generated questions span several domains and exhibit varied surface forms. These examples provide qualitative evidence of topical alignment. Gradient residual alignment measures whether a generated example provides a complementary optimization signal. Developing an explicit semantic or correctness filter for generated candidates is an important direction for improving this component.

\section{Limitations}
\label{sec:limitations}

The quality of the gradient similarity graph depends on the fidelity of the warmup checkpoints and the projected gradient features.
Our ablations show that QSGS is robust to a broad range of projection dimensions and trade-off parameters, yet very small warmup sets can produce noisy influence estimates.
In addition, our empirical study focuses on 7B--8B open-weight models and three representative benchmarks.
Extending the evaluation to larger models, additional domains, and multi-turn or tool-use tasks is an important direction for future work, especially because the same stored gradient features can be reused to explore new target sets with limited additional cost.

\end{document}